\documentclass[conference]{IEEEtran}
\IEEEoverridecommandlockouts
\usepackage{cite}
\usepackage{amsmath,amssymb,amsfonts}
\usepackage{algorithmic}
\usepackage{graphicx}
\usepackage{textcomp}
\usepackage{xcolor}
\def\BibTeX{{\rm B\kern-.05em{\sc i\kern-.025em b}\kern-.08em
    T\kern-.1667em\lower.7ex\hbox{E}\kern-.125emX}}
\usepackage{caption}
\usepackage{subcaption}

\begin{document}

\title{Time Series Anomaly Detection for Smart Grids: \\A Survey
}

\author{\IEEEauthorblockN{Jiuqi (Elise) Zhang, Di Wu, Benoit Boulet}
\IEEEauthorblockA{\textit{Department of Electrical and Computer Engineering} \\
\textit{McGill University}\\
Montreal, Canada \\
elise.zhang@mail.mcgill.ca, di.wu5@mail.mcgill.ca, benoit.boulet@mcgill.ca}
}

\maketitle

\begin{abstract}
With the rapid increase in the integration of renewable energy generation and the wide adoption of various electric appliances, power grids are now faced with more and more challenges. One prominent challenge is to implement efficient anomaly detection for different types of anomalous behaviors within power grids. These anomalous behaviors might be induced by unusual consumption patterns of the users, faulty grid infrastructures, outages, external cyberattacks, or energy fraud.
Identifying such anomalies is of critical importance for the reliable and efficient operation of modern power grids. Various methods have been proposed for anomaly detection on power grid time-series data. This paper presents a short survey of the recent advances in anomaly detection for power grid time-series data. 
Specifically, we first outline current research challenges in the power grid anomaly detection domain and further review the major anomaly detection approaches. Finally, we conclude the survey by identifying the potential directions for future research.

\end{abstract}


\section{Introduction}
The concept of smart grid refers to a power grid that can utilize smart electronic devices to obtain real-time monitoring of user actions and grid conditions, and make decisions towards more efficient energy management \cite{wu2017two, wu2018machine}. It has the potential to reduce peak demand \cite{Zhang2011Anomaly, wu2018optimizing}, improve energy conservation \cite{himeur2021artificial}, and enable the integration of renewable energy sources \cite{Pereira2018Unsupervised, wu2017boosting}, which sheds light on a smart and sustainable energy future. It has been receiving more and more attention with the fast rise in renewable energy generation and the increasing adoption of various electric appliances into the grid.

Many devices have been installed in modern power grids to further improve their operational efficiency. However, the installation of intelligent devices into the grid network has also rendered the system more susceptible to malfunctions and cyberattacks \cite{Gunduz2018Analysis, Skopik2014Dealing, Peter2017Cyber}. This can lead to utility fraud, data breach, energy theft, and potentially compromise the entire network, which can be extremely dangerous and costly. Identifying these 
anomalous behaviors requires the development of efficient anomaly detection techniques, which is critical to the enhancement of reliability and operational efficiency of grid networks. 

Most anomalous behaviors in power grids can be detected by analyzing time-series data. In a nutshell, all time series anomaly detection approaches follow the same fundamental idea - to characterize or forecast normal behaviors of time series and identify instances that deviate from the distribution of normal data points. Currently, many approaches have been proposed for electric load forecasting and anomaly detection, including classical time series analysis approaches \cite{contreras2003arima, chou2014real_ARIMA}, statistical approaches \cite{atsawathawichok2014long,Wei2020ESD} and machine learning-based approaches \cite{ceperic2013strategy, Smyl2020RNNForecast}. 

Time-series anomaly detection has been studied in various domains including economics \cite{hyndman2015large}, medicine \cite{chuah2007ecg}, etc. There have also been survey papers on general anomaly detection methodologies \cite{chalapathy2019deep}, time series anomaly detection for sensor faults \cite{gaddam2019anomaly, erhan2020smart} and Internet of Things (IoT) \cite{Cook2020AnomalyIoT}. However, there is no systematic review work yet for anomaly detection within the context of power grid. The purpose of this survey is to review recent advances of anomaly detection works on power grid time series.

The rest of this paper is organized as follows. In Section~\ref{Background}, we summarize the major types of anomalies that would be encountered in power grids and major research challenges. In Section~\ref{classical} and ~\ref{MachineLearning}, we discuss  classical anomaly detection approaches and machine learning-based approaches with applications to smart grid time-series data. In Sections~\ref{future} and~\ref{Conclusion}, we summarize the recent research findings and identify the potential future directions.

\section{Background}
\label{Background}

\subsection{Problem Background}
\label{anomalies}
Anomalies, or outliers, are data points that deviate from patterns or distributions of the majority of the data\cite{ruff2021unifying}. As presented in~\cite{Cook2020AnomalyIoT, ruff2021unifying}, there are three major types of anomalies for time-series data:

\begin{enumerate}
    \item \textit{Point Anomaly:} Point anomaly is one or a few observations that stand out from the range of all the other observations in the time series;

    \item \textit{Group Anomaly or Pattern Anomaly:} Group anomaly or pattern anomaly is a group or sequence of observations that deviate from the overall pattern or periodicity of the sequence;

    \item \textit{Contextual Anomaly:} Contextual anomaly is one or a sequence of observations that seems to be reasonable since it is within the range of observations in the sequence, but deviates from the "context", i.e., the pattern of its adjacent time range;
\end{enumerate}

\begin{figure*}[!h!]
\centering
\makebox[\textwidth][c]{
    \begin{subfigure}{0.36\textwidth}
    \centering
    \includegraphics[width=\linewidth]{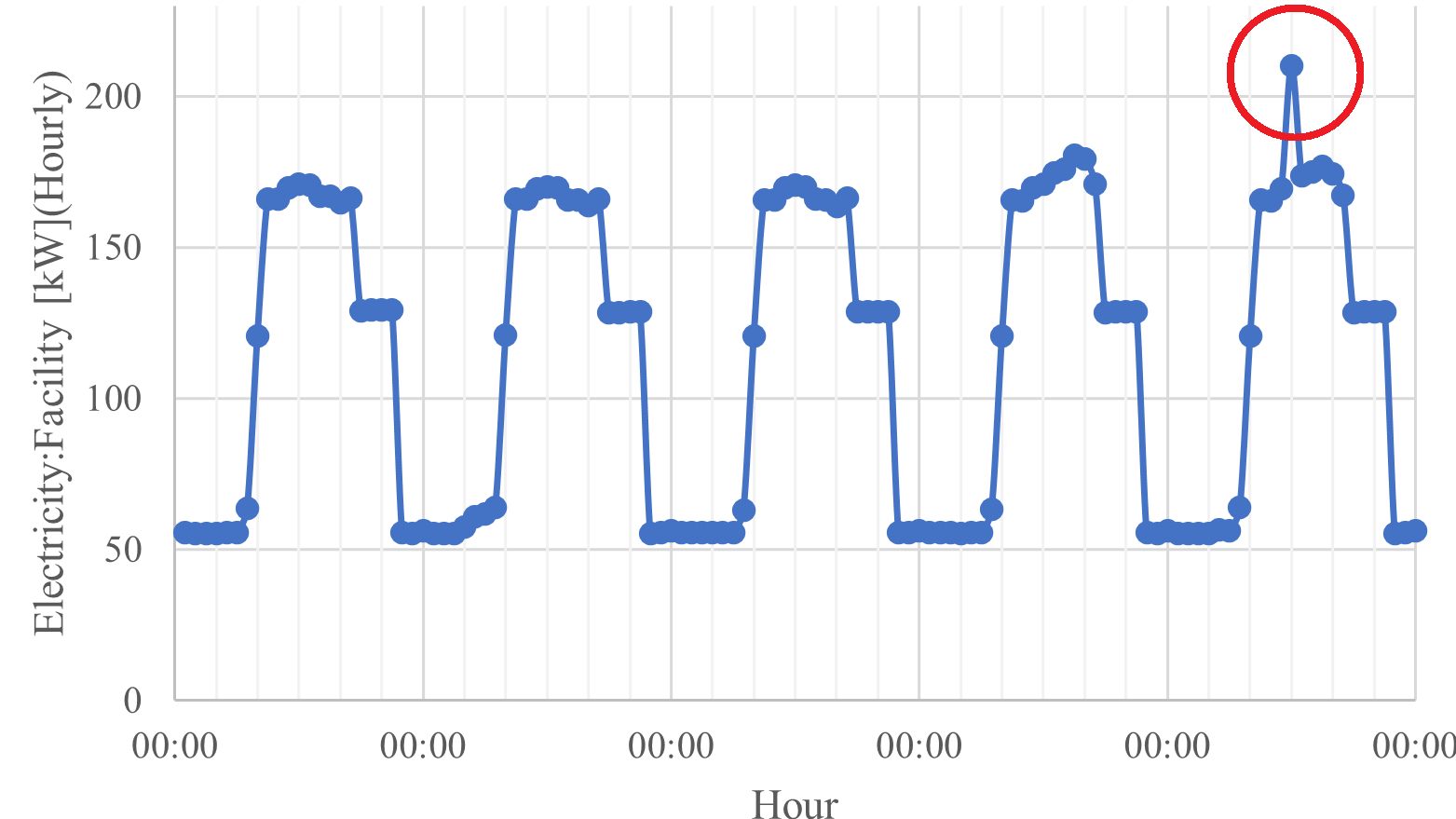}
    \caption{Point Anomaly}
    \label{fig:point_ano}
    \end{subfigure}
\hfill
    \begin{subfigure}{0.36\textwidth}
    \centering
    \includegraphics[width=\linewidth]{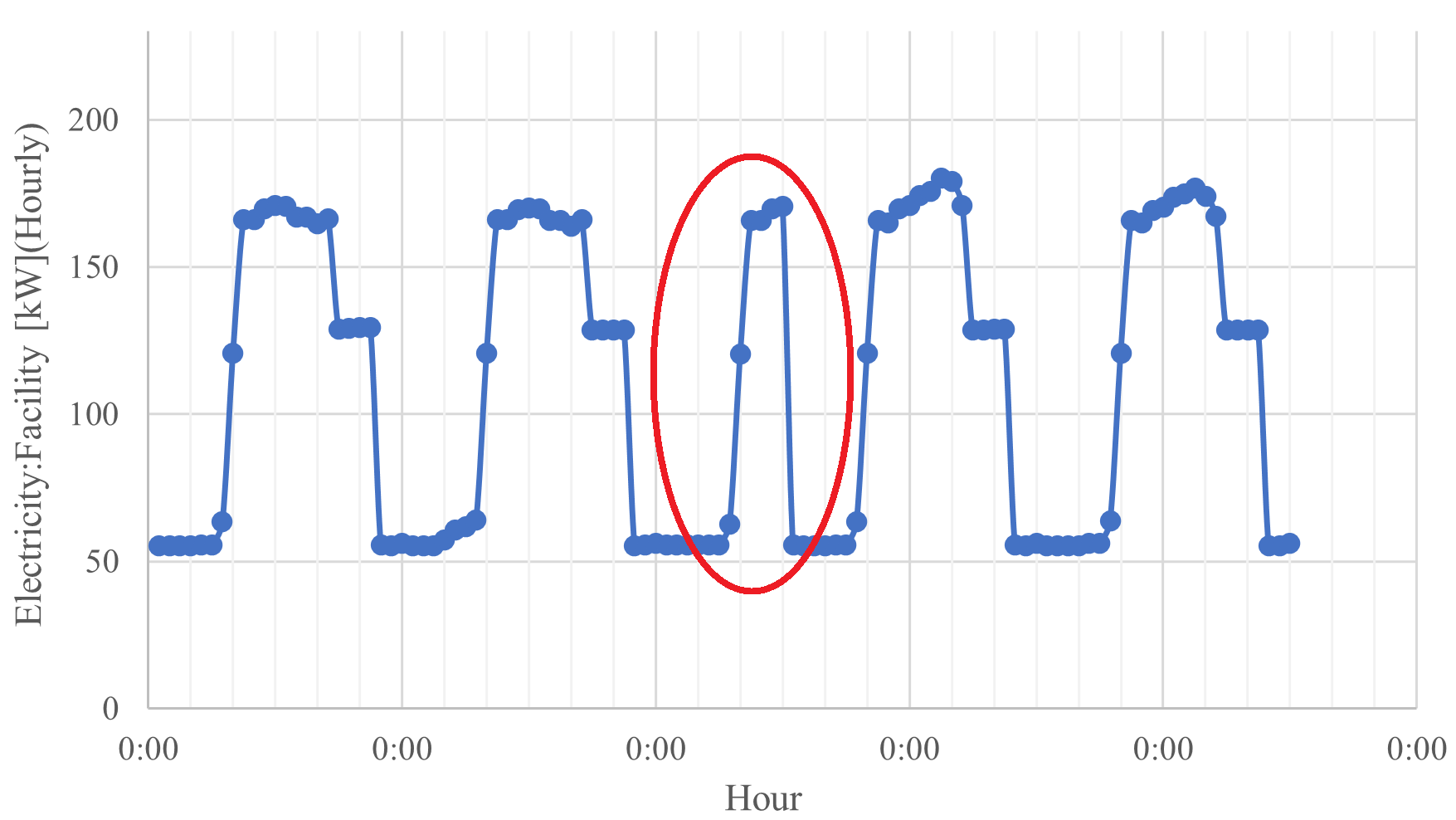}
    \caption{Pattern Anomaly}
    \label{fig:pattern_ano}    
    \end{subfigure}
\hfill    
    \begin{subfigure}{0.36\textwidth}
    \centering
    \includegraphics[width=\linewidth]{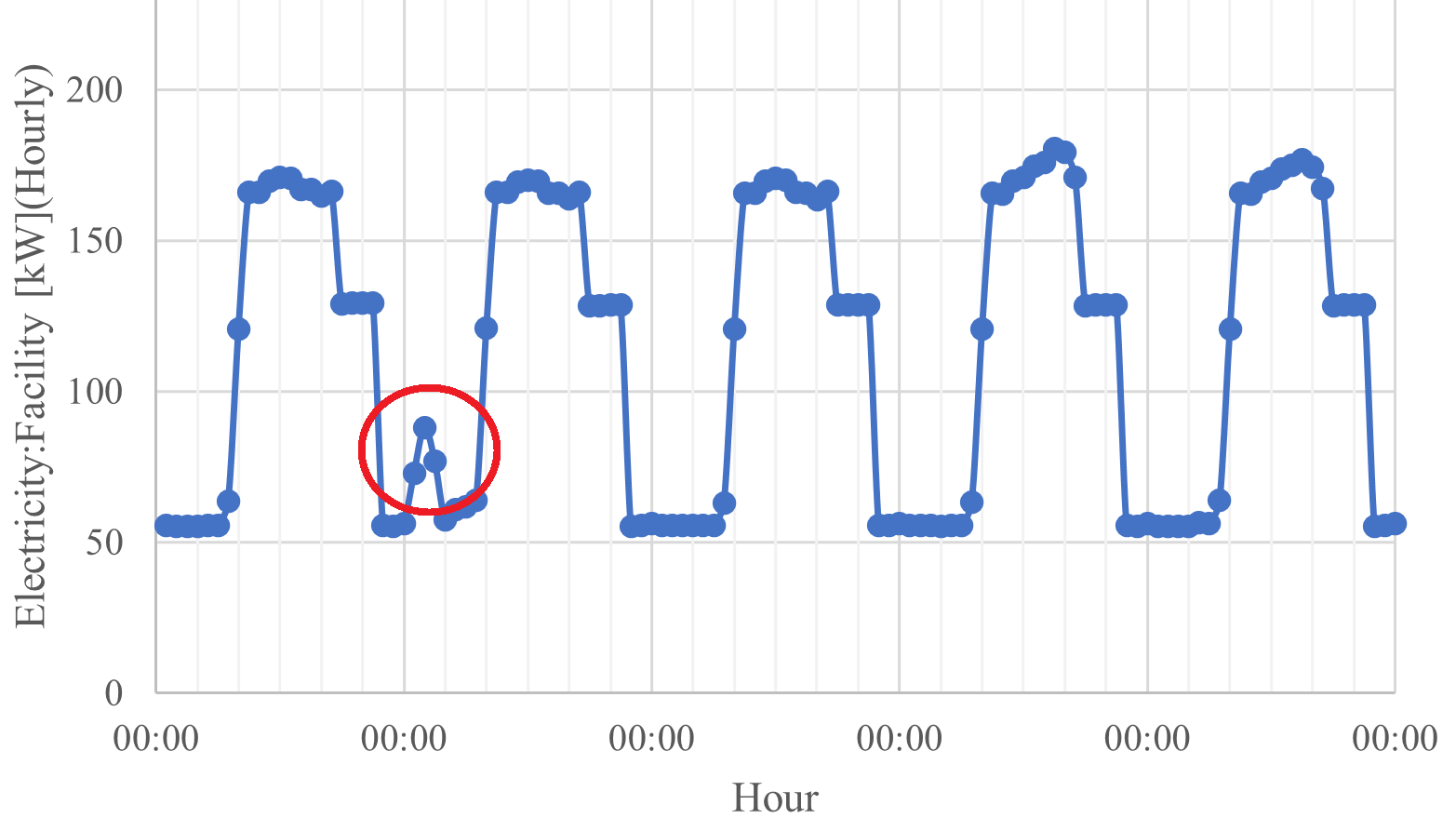}
    \caption{Contextual Anomaly}
    \label{fig:contextual_ano}    
    \end{subfigure}
}
\centering
\caption{Illustration of three types of outlier in an hourly load dataset of a restaurant. Data acquired from Open EI~\cite{OpenEI} }
\label{fig:commercial}
\end{figure*}

Following the above definition, anomalies within the power grid context can be defined as meter measurement data that represent unanticipated changes in either the generation, transmission, distribution, and consumption stages. 

Figure~\ref{fig:commercial} is an illustration of the three types of anomalies within an hourly electrical load profile of a commercial building. The normal data points follow a periodic pattern, whereas the anomalous data points in the red circles deviate from this pattern in three different ways. The point anomaly shown in Figure~\ref{fig:point_ano} is out of the range of all the other load measurements. The pattern anomaly shown in Figure~\ref{fig:pattern_ano} is the time series segment in which the periodicity of the load sequence changes. The contextual anomaly shown in Figure~\ref{fig:contextual_ano} are the data points whose values seem to be within a reasonable range according to the overall sequence while deviating from their adjacent data points and are therefore deemed to be outliers according to the context.


Most of works in this field are focused on the consumption process, specifically, to identify anomalous behaviors in power load or usage patterns. Other studies have also worked with datasets of phase measurement unit (PMU) measurements \cite{Song2017PowerCast}, real-time simulations on IEEE Bus testing platforms \cite{Karimipour2019ADeep}, power utility logs \cite{Skopik2014Dealing} and information exchanges between smart meters and micro-grids, e.g., grid communication protocol field \cite{Zhang2019Anomaly}, grid communication network traffic \cite{Feng2017MultiLevel}.

\subsection{Challenges}
\label{challenges}
There are several challenges hindering the effective anomaly detection in power grids.

\subsubsection{Limited Amount of Data}
In reality, there is a significant lack of labeled data. Most of the power grid time series are unlabeled - there are no clear indications on whether a specific data point is normal or anomalous, which poses challenges on the implementation of classification-based anomaly detection methods. Moreover, the imbalance between normal data and anomalies also makes it difficult to thoroughly capture the characteristics of anomalous data points.

\subsubsection{Online, Real-Time Detection}
Most existing anomaly detection models are offline. For these works, a sufficiently large amount of data is necessary for learning a reliable machine learning-based anomaly detection model.  Nevertheless, it may be extremely costly or simply impossible to collect enough data in reality (e.g., we may only have a very limited amount of data for the power load of newly constructed buildings). 

A more data-efficient anomaly detection should be online and in real-time, in which the model can utilize the current data at hand to make immediate, real-time predictions, and adapt itself to improve its performance over time \cite{liu2016regression}.

\subsubsection{Stationarity}
One fundamental assumption for most machine learning models is the stationarity of data distribution. Nevertheless, real-world time series are often non-stationary. Consider the constantly evolving power grid --- for the past ten years, renewable energy generation has been growing exponentially. Such changes would vary the distribution of power grid data and impose challenges on efficient and accurate anomaly detection.

\subsubsection{Noisy Inputs and Disturbance}
Erroneous meter measurements and various other types of uncertainties might inject noise or disturbance into power grids. This can come from the generation side, the power transmission, or the consumption side. For example, sudden changes in weather conditions might bring drastic changes to solar energy generation; power consumption might be significantly affected by changes in human behaviors; the sensitivity limit of grid meters might cause measurement errors.

\section{Classical Methods}\label{classical}
\subsection{Generalized Extreme Studentized Deviation Test}

The Generalized Extreme Studentized Deviation (ESD) Test \cite{ESD_Definition}, also known as the Grubbs's test \cite{grubbs1950sample}, is used for detecting outliers in univariate time series that approximately follow a normal distribution. For $X=\{x_1, x_2, x_3, ..., x_n\}$ and an upper bound $r$ for the estimated number of outliers, the residual is $R_{i}=\frac{\max _{i}\left|x_{i}-\bar{x}\right|}{s}$, where $\bar{x}$ and $s$ are the mean and standard deviation. Remove the $x_i$ that maximizes $\left|x_{i}-\bar{x}\right|$ and recompute the above residual with the remaining observations. Repeat for $i=1,2,3 ...,r$ until $r$ observations have been removed and obtain $r$ residuals, $R_1, R_2, ..., R_r$. The $r$ corresponding critical values based on the Student's $t$-Distribution are $\lambda_{i}=\frac{(n-i) t_{p, n-i-1}}{\sqrt{\left(n-i-1+t_{p, n-i-1}^{2}\right)(n-i+1)}}$, for $i=1,2,...,r$. Here $p=1-\frac{\alpha}{2(n-i+1)}$ ($\alpha$ is the significance level), $t_{p,v}$ is the $100p$ percentage point from the $t$-distribution with $v$ degrees of freedom. The largest $i$ with $R_i$ exceeding the critical $\lambda_i$ then indicates the number of outlier data points.

The performance of load forecasting models often depends on the quality of historical data. By adopting the ESD test to detect anomalies within the historical training data, one can follow the scheme of Random Sample Consensus (RANSAC) algorithm \cite{fischler1981random}, in which the outliers shall be detected by the ESD test and be removed in the preprocessing step, in order to improve the quality of the training data. In one application, the authors \cite{li2009classification} introduced the ESD test into their work to classify normal and abnormal energy consumptions and then removed the detected anomalies from the historical data for further modeling and forecasting. The authors of \cite{fan2014development} also used the ESD test to identify and remove anomalies within the energy consumption dataset of a commercial building in Hong Kong, and then proposed an ensemble method that ultimately achieved good performance on the next-day consumption prediction.

Another application of the ESD test for microgrid anomaly detection was investigated in \cite{Wei2020ESD}. A microgrid is a self-sufficient, decentralized electricity system. 
Using the Extreme Studentized Deviate test, the authors successfully identified and visualized the anomalies within a power load profile of a microgrid. One limitation of this approach is that it requires prior knowledge of the approximate number of outliers within the sequence since we need to provide an upper bound for computing the evaluation statistics.

\subsection{Autoregressive Moving Average-Based Methods} 

This family of models detects anomalies based on one-step-ahead prediction. For a time series $Y=\{ y_1,y_2,…,y_t \}$, an autoregressive moving average (ARMA) model is
$y_{t}=a_{1} y_{t-1}+a_{2} y_{t-2}+\ldots a_{p} y_{t-p}+n_{t}+b_{1} n_{t-1}+\cdots+b_{q} n_{t-q}$, where $n_i$ represents the error at time step $i$. Here $p$ is the total number of autoregressive terms, $q$ is the number of lagged errors, and $b$ is the moving average coefficients. An autoregressive integrated moving average (ARIMA) model further accounts for the non-stationary characteristics of certain types of time series and makes the sequence stationary in an initial difference step, in other words, eliminating the trend by subtracting previous values from current values. 

Anomaly detection of a load consumption profile of office space was studied in \cite{chou2014real_ARIMA}. The authors compared the standard ARIMA model and the neural network (NN)-based ARIMA model in terms of their predictive ability and anomaly detection performance. Since the standard ARIMA model tends to converge towards the mean value in long-term predictions, it failed to capture consumption patterns and was not able to distinguish between normal and anomalous consumption under the two-sigma rule. On the other hand, the non-linear activation layers in the neural network (NN)-based ARIMA enabled the model to capture the long-term consumption pattern and achieved satisfactory prediction accuracy in an 8-week power consumption time series.


\section{Machine Learning-Based Methods}
\label{MachineLearning}

This section elaborates on the machine learning approaches for power grid anomaly detection. 

\subsection{Supervised Classification-Based Methods}
\label{supervised}
Anomaly detection can be considered as a classification task in specific scenarios. Supervised classification-based methods require the training set to have labels. One study \cite{Esmalifalak2017Detecting} worked on detecting false data injection in power flow measurements of IEEE 118-bus test system using a Gaussian kernel support vector machine (SVM). In this case, the dataset is labeled, since the false data injections were simulated by the authors.  
Another study \cite{himeur2020novel} on detecting anomalous consumption behaviors adopted the idea of Google’s micro-moments. It used a rule-based model to assign ``micro-moment'' labels to the energy consumption data (``normal'' labels include \textit{``good usage”, ``turn on device”} and \textit{``turn off device”}; ``abnormal'' labels include \textit{``excessive consumption”} and \textit{``consumption while outside”}), which transforms the anomaly detection problem into a classification problem. The authors utilized a deep neural network classifier and achieved high detection accuracy. 

\subsection{Unsupervised Methods}
On the other hand, unsupervised methods do not need class labels but seek to capture intrinsic patterns of the data itself. This type of techniques is based on the assumption that normal data occurs much more frequently than abnormal data \cite{Chandola2009AnomalySurvey}. The essence is to train a model to predict the normal behavior of the time series, and to flag as anomalies that are unlikely to be sampled from the forecasted distribution. Famous techniques include reconstruction-based methods, clustering-based methods, and prediction-based methods.

\subsubsection{Reconstruction-Based Methods}
\label{Reconstruction}
Reconstruction-based anomaly detection is a process of feature extraction or dimensionality reduction, by which an initial set of raw data is reduced to a smaller but more meaningful set of features ~\cite{khalid2014survey}. During the training stage, reconstruction-based models look to extract information of historical patterns from the original time series. In the testing stage, when a new observation comes in, the features of the new data instance are compared with extracted historical patterns. The further it deviates from the patterns, the more likely it will be flagged as an anomaly. 

Feature reconstruction can either facilitate the preprocessing and labeling of raw training data and be further applied to unsupervised anomaly detection or can be used directly as an anomaly detection scheme. 

In a study on cyberattack detection for large-scale Smart Grids \cite{Karimipour2019ADeep}, Symbolic Dynamic Filtering (SDF) is used as a feature extraction strategy. The phase space of the original time series is partitioned into a finite number of cells. Then, by assigning a symbol to each partition, the time series is compressed to a symbolic sequence. This facilitates the later implementation of Dynamic Bayesian Network (DBN) for revealing the causal relationship between these symbolic features and Restricted Boltzmann Machine (RBM) for capturing the distribution of normal operation of the grid system.

Principal component analysis (PCA) is a most popular method for feature extraction. It produces a reduced embedding of approximately independent features based on the covariance matrix of time series. In \cite{imayakumar2020anomaly_PCA}, the authors used PCA-based methods as a direct approach for anomaly detection on primary distribution voltage magnitude measurements. The measurement time series was first converted to a matrix representation, followed by PCA to extract independent features and reduce the raw data to low-dimensional projection space.  Anomaly detection was conducted based on the residual between the new observation and the projection space. Another descriptive symbolic feature extraction method, Symbolic Aggregate ApproXimation (SAX), was proposed for power load anomaly detection in \cite{Yue2017Integrated, Yue2019Descriptive}. In a SAX implementation, the time series is first segmented into intervals, then a symbol is assigned to each segment based on the mean value. In this way, time series is converted to a low dimensional, discrete representation for further processing and anomaly detection.

\subsubsection{Clustering-Based Methods}
\label{clustering}
Clustering-based method models the underlying characteristics of data by projecting data instances to a higher dimensional space and formulating clusters based on distance metrics \cite{Chandola2009AnomalySurvey,Cook2020AnomalyIoT}. Data points closer to dense clusters belong to the majority and are more likely to be normal observations, while data points farther away from the clusters are usually flagged as anomalies.

K-means clustering algorithm \cite{hartigan1979algorithm,jain2010data} is one popular unsupervised clustering method. It is simple and efficient - given prior knowledge of the number of clusters $k$, the algorithm adjusts the location of $k$ cluster centroids in an iterative manner based on the distance between data instances and centroids. In \cite{fenza2019drift}, k-means clustering was applied on the multivariate time series of the electricity consumption of 370 users and formulated a set of preliminary class labels for each user's consumption. This result would facilitate the consumption pattern drift detection in later steps. 

 Isolation Forest (iForest) \cite{IsolationForest} is another efficient unsupervised anomaly detection approach and can be used for clustering. In this method, each isolation tree partitions the instance space recursively, and then we compute the average path length from each data instance to all the tree roots. Based on the assumption that anomalous data points are few and distinct from normal data, it should be relatively easy to distinguish anomalies from the normal data, as a result, the anomalies should be isolated by the partition trees within a few steps. Hence, normal data points are usually located in deeper branches of the trees, while outliers are located much closer to the root. A shorter path length to the root then indicates a higher probability of being an anomaly. In \cite{mao2018anomaly_IsoForest}, the authors proposed a pattern-based outlier detection approach based on Isolation Forest. Power consumption time series is discretized using a fixed-length sliding window. Next, the mean, standard deviation, and trend of each segment are computed as raw features, followed by PCA for feature space dimensionality reduction. Isolation forest is then applied to detect anomalies within the reduced dataset and achieved high detection accuracy.

\subsubsection{Prediction-Based Methods}
Prediction-based methods rely on accurate forecasts of future time steps. Recurrent Neural Networks (RNNs) are a family of models especially suitable for handling sequential data. Long-short-term-memory (LSTM) RNN \cite{Hochreiter1997lstm} is an RNN architecture with input, output, and forget gates that help resolve the vanishing gradient problem of a vanilla RNN. It has found applications in various time series forecasting tasks and has demonstrated promising performance \cite{Connor1994RNNForecast, Zhang2000RNNForecast, Assaad2008RNNForecast, Smyl2020RNNForecast}.


An RNN-based anomaly detection method was proposed for the power load data in~\cite{fenza2019drift}. LSTM-RNN produces a one-step-ahead prediction of power load. Then, the prediction error is computed and compared with the standard deviation of all prediction errors. The performance evaluation is based on the two-sigma rule from the property of a normal distribution, where roughly 95\% of data lie within the confidence range of $(-2\sigma,2\sigma)$. Precision and recall were then adopted as the evaluation metrics for the model. Other studies on RNN-based models for power grid anomaly detection followed a similar idea, but have adopted different evaluation metrics, including Manhattan distance and edit distance \cite{fengming2017anomaly}, autoencoder reconstruction score \cite{Pereira2018Unsupervised} and absolute relative error \cite{nguyen2018applications}.

\subsection{Ensemble Methods}

By utilizing several base models jointly, ensemble methods can help improve the performance of base learners~\cite{polikar2012ensemble}. The efficient operation of power delivery systems relies on accurate load forecasts. The anomaly detection of historical load data is tackled in the following studies \cite{Yue2017Integrated, Yue2019Descriptive}. The authors proposed an integrated solution composed of three anomaly detectors for power consumption data, namely, a Chebyshev Inequality (CI)-based model, a Second Order Difference (SOD)-based model, and Symbolic Aggregation approXimation (SAX). Each base model computes an anomaly metric, and the final result is obtained by leveraging the prediction of all three detectors. 

Another study \cite{Zhou2019Ensemble} worked with synchrophasor data collected by phased measurement units (PMUs). The authors adopted an unsupervised ensemble model composed of three base detectors, namely, a Chebyshev Inequality (CI)-based detector, a Linear Regression (LR)-based detector, and a Density-Based Spatial Clustering of Applications with Noise (DBSCAN) clustering detector. Anomaly scores generated by the base detector are going to facilitate the final decision-making of the ensemble model.

\section{Future Research Directions}
\label{future}

Although time series anomaly detection has been studied for years, its implementation still faces domain-specific challenges with the continuing emergence of new application fields. We hereby highlight three potential directions for future research.

\subsection{Building Anomaly Benchmarks for Smart Grid Time Series}

The lack of sufficient anomalous data in smart grid time series hinders characterization of the anomalous class. In this regard, building anomaly benchmarks for power grid time series would be beneficial for the development of accurate anomaly detection models. 

Here we propose four strategies to follow when building anomaly benchmarks: 

\begin{enumerate}
    \item \textit{Manual Labeling:} Flag anomalies manually in unlabeled raw datasets. This might be the most ideal scenario but is extremely time and resource-consuming.
    \item \textit{Anomalous Classes:} Assign class labels to each data instance based on preliminary clustering results and convert the problem to a supervised classification problem by selecting one or several classes to be anomalous.
    \item \textit{Synthetic Anomalies:} Generate synthetic anomalies either based on prior knowledge of the statistical distribution of anomalies or by simulating disturbances or attacks on the original data. 
    \item \textit{Adversarial Attacks:} Anomalies are maliciously designed based on prior knowledge of the detection model in order to deceive the detection algorithm.
\end{enumerate}

Depending on the scenario we are dealing with, appropriate strategies shall be utilized to build corresponding anomaly benchmarks.

\subsection{Transfer Learning in Anomaly Detection}

Transfer learning~\cite{wu2017boosting, wu2019multiple} aims to improve the learning performance on the target domain via reusing the knowledge learned from source domains. It has recently been implemented in tasks including anomaly detection for surveillance videos \cite{bansod2019transfer}, image classification \cite{andrews2016transfer}, time series classification \cite{fawaz2018transfer} and time series forecasting \cite{ye2018novel}. In the future, it would also be interesting to explore the potential benefit of transfer learning for power grid anomaly detection, which has the potential for improving the model generalization in the face of new environments. For newly-constructed appliances or buildings where sufficient historical data is not yet available, we can utilize prior knowledge extracted from existing labeled datasets, or from the historical data of similar appliances or buildings. 

\subsection{Data-Efficient Online Anomaly Detection in Smart Grids}

Most current models are trained offline using a large amount of historical data and might not be applicable for real-time online anomaly detection. Real-time detection models suffer from a lack of data in the beginning stages when the meter measurements are just starting to be collected. Hence, it would be interesting to investigate how to effectively utilize all available data at hand.

Generative models are capable of directly capturing the underlying distribution of data and generate new instances, which makes it a promising approach towards our goal. In recent years, Generative Adversarial Network (GAN)-based models have demonstrated excellent performance on tasks including realistic time-series imputation \cite{luo2018multivariate}, time series generation \cite{esteban2017realvalued} and power load forecasting \cite{zhang2020deep}. Following this idea, GAN-based models might be used as a real-time data augmentation scheme. Synthetic data can then be generated in real-time to augment the current input data.

\section{Conclusion}
\label{Conclusion}
In this survey, we summarize the definition of anomalies in time-series data and identify the major research challenges within the smart grid context. We then discuss the classical and machine learning-based techniques for anomaly detection on power grid time series and propose three potential future research directions. In conclusion, we believe that anomaly detection would continue to play a crucial role in the future quest for the reliability and operational efficiency of power grids. As recent advances in ensemble learning, transfer learning, and generative modeling present new opportunities to resolve current challenges, we would expect anomaly detection models to be more refined and accurate towards applications in smart grid time series.

\bibliographystyle{IEEEtran}
\bibliography{ref.bib}

\end{document}